\documentclass{article}
\usepackage{spconf,amsmath,graphicx}

\usepackage{enumitem}
\setlist{nosep, leftmargin=*}
\usepackage[table]{xcolor}
\usepackage{amssymb, amsfonts}
\usepackage{euscript}
\usepackage{pifont}
\usepackage{caption}
\usepackage{hyperref}
\usepackage{cleveref}
\usepackage{mathtools}
\usepackage{booktabs}
\usepackage{multirow}
\usepackage{threeparttable}
\usepackage{subcaption}
\graphicspath{Figures/}
\usepackage{graphicx}
\usepackage[export]{adjustbox}

\title{FuseNet: Self-Supervised Dual-Path Network for Medical Image Segmentation}

\name{\begin{tabular}{@{}c@{}}
Amirhossein Kazerouni*$^{1}$\thanks{$^*$Equal contribution. This work is partially supported by NIH R01-CA246704, R01-CA240639, R03-EB032943, U01- DK127384-02S1, and U01-CA268808.} \qquad 
Sanaz Karimijafarbigloo*$^{2}$ \qquad 
Reza Azad*$^{3}$ \qquad 
Yury Velichko$^{4}$ \\ 
Ulas Bagci$^{4}$ \qquad 
Dorit Merhof$^{2,5}$
\end{tabular}}

\address{$^{1}$School of Electrical Engineering, Iran University of Science and Technology, Iran 
\\
 $^{2}$ Faculty of Informatics and Data Science, University of Regensburg, Germany 
 \\
 $^{3}$ Faculty of Electrical Engineering and Information Technology, RWTH Aachen University, Germany
 \\
 $^{4}$Department of Radiology, Northwestern University, Chicago, IL, USA
 \\
 $^{5}$Fraunhofer Institute for Digital Medicine MEVIS, Germany
}

\begin{document}

\maketitle
\begin{abstract}
Semantic segmentation, a crucial task in computer vision, often relies on labor-intensive and costly annotated datasets for training. In response to this challenge, we introduce FuseNet, a dual-stream framework for self-supervised semantic segmentation that eliminates the need for manual annotation. FuseNet leverages the shared semantic dependencies between the original and augmented images to create a clustering space, effectively assigning pixels to semantically related clusters, and ultimately generating the segmentation map. Additionally, FuseNet incorporates a cross-modal fusion technique that extends the principles of CLIP by replacing textual data with augmented images. This approach enables the model to learn complex visual representations, enhancing robustness against variations similar to CLIP's text invariance. To further improve edge alignment and spatial consistency between neighboring pixels, we introduce an edge refinement loss. This loss function considers edge information to enhance spatial coherence, facilitating the grouping of nearby pixels with similar visual features. Extensive experiments on skin lesion and lung segmentation datasets demonstrate the effectiveness of our method. \href{https://github.com/xmindflow/FuseNet}{Codebase.}
\end{abstract}
\vspace{-0.25em}
\section{Introduction}
\vspace{-0.75em}
\label{sec:intro}
Recently, deep learning methods have achieved cutting-edge results in various computer vision tasks, including semantic segmentation~\cite{azad2022medical}. However, training deep learning models typically requires large and well curated annotated datasets due to the millions of trainable parameters involved. The collection of such training data primarily relies on manual annotation, which is often quite costly, especially in the context of semantic segmentation, due to the need for pixel-level precision.
Recently, self-supervised learning has emerged as a promising alternative to traditional well-curated labeled data. By extracting meaningful representations from the inherent structure of unlabeled data, it eliminates the reliance on supervised loss functions that require manual annotations. This approach has demonstrated success in a variety of medical imaging applications, including dermatological imaging~\cite{gharawi2023self} and radiology scans~\cite{karimijafarbigloo2023ms}, among others. It is also important to acknowledge that semantic segmentation is intricately interwoven with local texture and global image context dependencies. Numerous studies have shown that simultaneous learning of local and global representations can significantly improve the accuracy of dense predictions~\cite{azad2022medical,gharawi2023self,karimijafarbigloo2023self}. 
In a related study, Ahn et al.~\cite{ahn2021spatial} introduced the SGSCN network, which uses multiple loss functions to group spatially connected pixels with similar features, enabling an iterative learning of pixel features and clustering assignments from a single image. Taher et al.~\cite{taher2022caid} developed the Context-Aware instance Discrimination (CAiD) framework to improve instance discrimination learning in medical images. CAiD extracts detailed and discriminative information from different local contexts in unlabeled medical images. Karimi et al.~\cite{karimijafarbigloo2023ms} presented a dual-branch transformer network that captures both global context and local details. This network utilizes self-supervised learning by considering semantic relationships between different scales, ensuring inter-scale consistency, and enforcing spatial stability within each scale for self-supervised content clustering. Another approach~\cite{gharawi2023self} aimed to address the lack of local and boundary representations by combining the CNN and vision transformer features. He et al.~\cite{he2023geometric} introduced Geometric Visual Similarity Learning, a method that incorporates topological invariance to measure inter-image similarity and create consistent representations of semantic regions.

\begin{figure*}[!ht]
\centering
\begin{subfigure}{0.67\textwidth}
    \includegraphics[width=\textwidth]{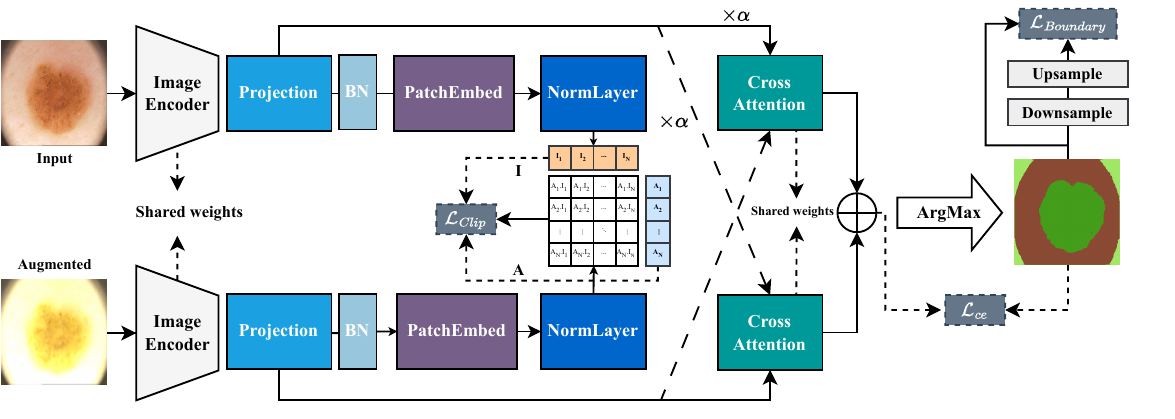}
    \label{fig:overview}
\end{subfigure}
\hfill\rule[5pt]{1pt}{145pt}\hfill
\begin{subfigure}{0.26\textwidth}
    \includegraphics[width=\textwidth]{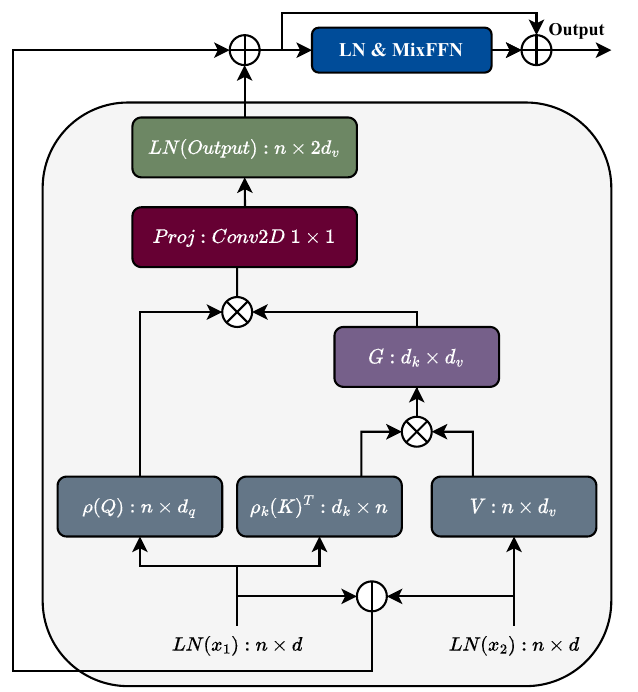}
    \label{fig:cross}
\end{subfigure}
\vspace{-1em}
\caption{A general overview of the proposed FuseNet (left) and the cross-attention module (right).}
\vspace{-1.5em}
\label{fig:mainfig}
\end{figure*}
Motivated by the existing research in this field, we aim to address the following question: \textit{How can we leverage self-supervised techniques to model local and global consistency and enhance semantic segmentation?} To address this challenge, we introduce \textit{FuseNet}, a novel self-supervised approach to semantic segmentation that aims to achieve a balance between local texture details and global context dependencies. \textbf{Our contributions include: }\ding{202}
A self-supervised approach for semantic segmentation that minimizes the reliance on expensive  manual annotations.
\ding{203} Integration of self-supervised learning to simultaneously capture local and global image characteristics.
\ding{204} Introduction of cross-modal fusion, which enhances the model's ability to handle complex scenarios.
\ding{205} Improved edge alignment and spatial consistency between adjacent pixels.

\vspace{-1.25em}
\section{Proposed Method}
\vspace{-0.75em}
We present \textbf{FuseNet} (see \autoref{fig:mainfig}), a dual-stream, self-supervised approach for image segmentation, which obviates the need for manual annotations. In FuseNet, while one stream processes the original image, the other handles its augmented counterpart, fostering data diversity, and enhancing robustness, invariance, and segmentation quality for real-world scenarios. Crucially, our framework facilitates the exchange of information between these two pathways before finally fusing their insights, which contributes to the model's enhanced performance. Our approach also incorporates data-driven loss functions, facilitating effective content clustering.
\vspace{-1em}
\vspace{-1.4em}
\subsection{Network Architecture}
\vspace{-0.5em}
Dual-path architectures have proven their effectiveness in self-supervised learning by leveraging the benefits of dual data views, leading to reduced overfitting and enhanced generalization~\cite{karimijafarbigloo2023ms}. They employ both the original and transformed data to enrich feature representations, fostering robustness. Inspired by this architecture, we have developed a dual-stream framework that simultaneously processes the original and augmented data, thereby enhancing adaptability and resilience across diverse scenarios. First, we apply data augmentation to the image ($ \in \mathcal{R}^{H \times W \times C}$), utilizing techniques such as ColorJitter and GaussianBlur, to create an augmented version of the image. This process is effective because it introduces controlled variations, i.e. color changes and blurring, which help the model learn to be invariant to different transformations while preserving the overall structural and semantic information. 
Notably, we refrain from employing transformation augmentation techniques that could potentially compromise the quality of the outcomes.
Subsequently, both the original and augmented images are then simultaneously fed into a shared weight encoder, which consists of a $3\times3$ convolutional layer followed by batch normalization (BN), a $1\times1$ depth-wise convolution, and another BN layer. This straightforward architecture is designed to facilitate the embedding of the input images into a high-dimensional feature space, enabling the model to capture intricate and localized patterns within the data. The subsequent projection block plays a crucial role in disentangling and refining meaningful, invariant features within the input data. Post-projection batch normalization is employed to standardize feature distributions, effectively alleviating internal covariate shifts and enhancing network generalization and training stability. The projection block is as follows:
\vspace{-0.5em}
\begin{equation}
    \textbf{Projection} = \text{LN}(\text{Linear}_2(\text{GELU}(\text{Linear}_1(x)) + x),
\end{equation}
\noindent where LN denotes LayerNorm layer. Next, the PatchEmbed and NormLayer are utilized to segment the input features into smaller patches and normalize these patches, respectively. The normalization is achieved by dividing the patch features by their L2 norms. This process yields two tokenized sequences, denoted as $I \in \mathbb{R}^{(\frac{H}{p} \frac{W}{p}) \times p^{2}C}$ and $A \in \mathbb{R}^{(\frac{H}{p} \frac{W}{p}) \times p^{2}C}$, where $p$ represents the patch size set as $\frac{H}{8}$. These feature maps are then used for cross-modal fusion.

To facilitate effective information exchange between the image features and the augmented image features derived from the projection heads, we employ the cross-attention module as illustrated in \autoref{fig:mainfig}. This enhancement strengthens the model's ability to grasp and utilize the shared attributes and dissimilarities between the original and augmented views of the data, fostering a more comprehensive understanding of the data's distribution, cross-modal relationships, and local dependencies. To give greater emphasis to the input image features, we apply a coefficient weight $\alpha$ to scale these features. The cross-attention weight is also shared between the two streams. In the image stream, $x_1$ represents to the original image features and $x_2$ represents the augmented image features. In the augmented image stream, the roles of $x_1$ and $x_2$ are reversed. The detailed implementation of the cross-attention block is depicted in 
\vspace{-0.5em}
\autoref{eq:cross}:
\begin{align}
    &\textbf{Q}, \textbf{K} = \text{Proj}(\text{LN}(x_1)), \textbf{V} = \text{Proj}(\text{LN}(x_2)),\nonumber\\
    &\textbf{X} = [\text{LN}(x_1) || \text{LN}(x_2)],\nonumber\\
    &\textbf{E} = \rho_q(\textbf{Q})(\rho_k(\textbf{K})^{T}\textbf{V}),\nonumber \\
    &\textbf{T} = \textbf{X} + \text{LN}(\text{Conv}{1 \times 1}(\textbf{E})),\nonumber\\
    &\textbf{Output} = \textbf{T} + \text{MixFFN}(\text{LN}(\textbf{T})),\label{eq:cross}
\end{align}
where $proj$ refers to a linear projection layer, $\rho_k$ and $\rho_k$ are SoftMax normalization functions, and MixFFN is a feed-forward network adopted from~\cite{huang2021missformer}.

Finally, the outputs from both streams are combined by summation, resulting in a soft prediction map $P \in R^{ H\times W \times K}$, where $K$ represents the number of clusters. To obtain the final semantic segmentation map $Y$ of the same dimensions, we apply he ArgMax function to determine the cluster index for each spatial location. During training, our network iteratively minimizes the cross-entropy loss, which quantifies the discrepancy between the soft prediction map and the segmentation map.
\autoref{eq:ce} shows the cross-entropy loss in our framework:
\vspace{-0.5em}
\begin{equation}
\mathcal{L}_{ce}\left(\mathbf{P}, \mathbf{Y}\right)=-\frac{1}{H \times W} \sum_{i=1}^{H \times W} \sum_{j=1}^K \mathbf{Y}_{i, j} \log \left(\mathbf{P}_{i, j}\right).
\label{eq:ce}
\end{equation}

Our approach leverages cross-entropy loss to learn cluster distribution by bolstering the network's confidence in grouping similar pixels. However, it faces challenges in modeling local spatial relationships, which can impact performance in merging adjacent clusters. To address this limitation, we introduce two additional regularization terms: the cross-modal fusion loss and the edge refinement loss.
\vspace{-1em}
\subsection{Cross-Modal Fusion}
\vspace{-0.5em}
In addition to cross-entropy loss, we introduce a cross-modal fusion approach that enhances the integration of information from both original and augmented image data. This approach encourages the model to develop a unified understanding of both augmented images and their corresponding originals, fostering robust learning. Our approach extends the principles of CLIP~\cite{radford2021learning} by substituting textual data with augmented images, introducing novel advantages specific to our model. This adaptation enables our model to acquire intricate visual representations, effectively aligning with the complexity of the data at hand. The controlled variations introduced by augmentations promote robustness, similar to CLIP's invariance to textual variations, which is critical for real-world data with unpredictable transformations. The CLIP loss is as follows:
\vspace{-0.5em}
\begin{align}
    &\textbf{Logit} = (I \cdot A^T)/{T},\nonumber \\
    &\textbf{Target} = \text{SoftMax}(({I\cdot I^T + A\cdot A^T})/2T),\nonumber \\
    &\mathcal{L}_{CLIP}= \left ( {\mathcal{L}_{ce}(\text{Logit}, \text{Target})+\mathcal{L}_{ce}(\text{Logit}^{T},\text{Target}^{T})} \right )/{2},
\end{align}
\noindent where $T$ is a temperature parameter. The CLIP loss in our framework aims to align the feature representations of the original image with its augmented counterpart. This alignment strengthens the model's ability to comprehend the shared features and differences between these two perspectives of the same data.
\vspace{-1em}
\subsection{Edge Refinement}
\vspace{-0.5em}
To improve edge alignment and promote spatial consistency among adjacent pixels, we introduce the edge refinement loss. This loss function aims to minimize the discrepancy between the segmentation map and its downsampled and subsequently upsampled counterpart, which generates an edge map. By minimizing this loss, our edge refinement technique enhances spatial coherence, encouraging the grouping of neighboring pixels with similar visual features. This approach involves downsampling an image by a factor of $\beta$ and then upsampling it. This allows us to prioritize key objects within the image and then accurately delineate object boundaries by subtracting the upsampled image from the original segmentation map. As a result, this method leads to improved consistency in spatial relationships and more precise object boundary delineation. The edge refinement loss is defined as follows:
\vspace{-0.5em}
\begin{align}
\mathcal{L}_{\text{Boundary}} = \sum_{i,j} (\left| (\text{Down-Up-Y})_{i,j} - \text{Y}_{i,j} \right|,
\end{align}
where $(\text{Down-Up-Y})_{i,j}$ and $\text{Y}_{i,j}$, and shows the Downsampled-Upsampled-segmentation map and segmentation map at pixel location $(i,j)$, respectively.
\vspace{-1em}
\subsection{Joint Objective}
\vspace{-0.5em}
The final loss function used in our training process is a combination of three distinct loss terms, as outlined below:
\vspace{-0.5em}
\begin{equation} \label{losses} \mathcal{L}_{\text {joint}}=\lambda_1 \mathcal{L}_{ce}+\lambda_2 \mathcal{L}_{CLIP}+\lambda_3 \mathcal{L}_{Boundary},
\end{equation}
where in order to control the relative importance of each loss term, we introduce weighting factors $\lambda_1$, $\lambda_2$, and $\lambda_3$. 


\begin{figure*}[t]
    \centering
    \resizebox{\textwidth}{!}{
    \begin{tabular}{@{} *{12}c @{}}
    \includegraphics[width=0.20\textwidth]{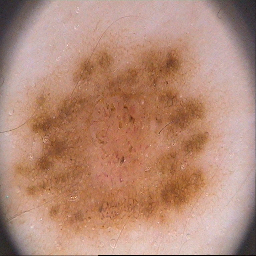} &
    \includegraphics[width=0.20\textwidth]{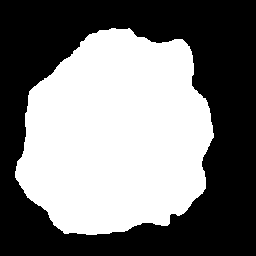} &
    \includegraphics[width=0.20\textwidth]{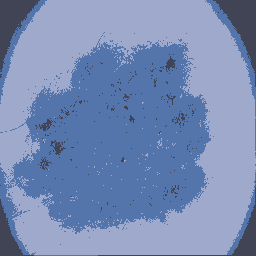} &
    \includegraphics[width=0.20\textwidth]{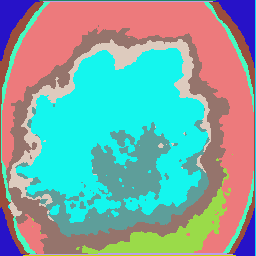} &
    \includegraphics[width=0.20\textwidth]{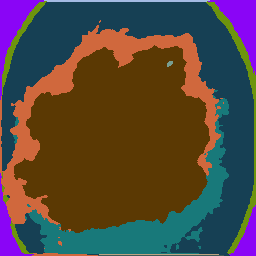} &
    \includegraphics[width=0.20\textwidth]{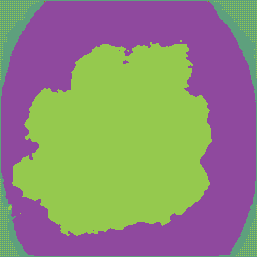} &
    \includegraphics[width=0.20\textwidth]{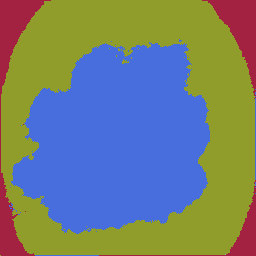} &
    \includegraphics[width=0.3cm, height=3.5cm]{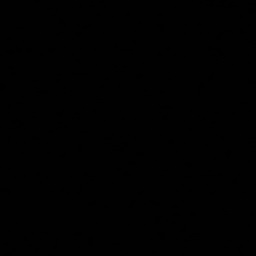} &
    \includegraphics[width=0.20\linewidth]{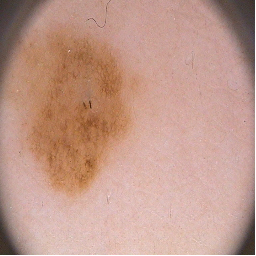} &
    \includegraphics[width=0.20\linewidth]{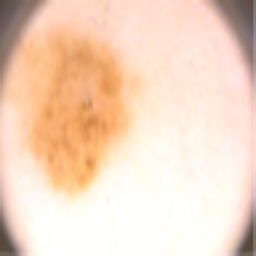} &
    \includegraphics[width=0.20\linewidth]{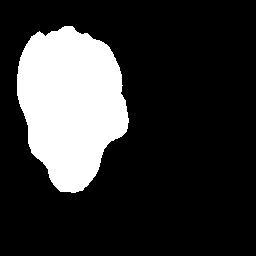} &
    \includegraphics[width=0.20\linewidth]{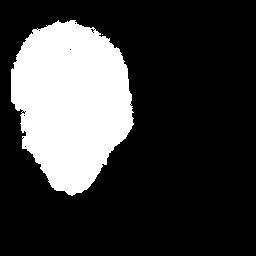}
    \\
    & & & & & & & \includegraphics[width=0.3cm, height=0.5cm]{Figures/dark_patch.jpg} & {\Large Input Image} & {\Large Aug. Image} & {\Large Ground Truth} & {\Large Our Method} 
    \\
    \includegraphics[width=0.20\textwidth]{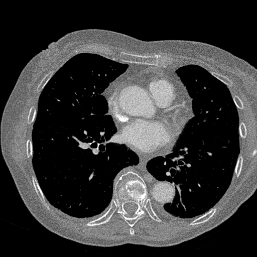} &
    \includegraphics[width=0.20\textwidth]{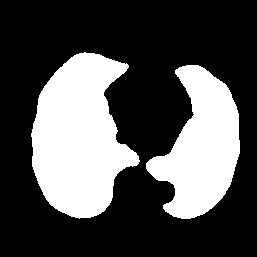} &
    \includegraphics[width=0.20\textwidth]{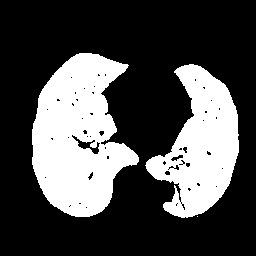} &
    \includegraphics[width=0.20\textwidth]{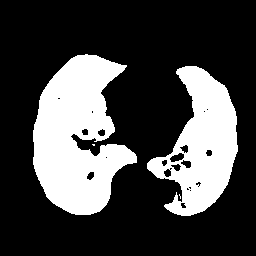} &
    \includegraphics[width=0.20\textwidth]{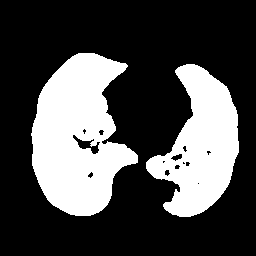} &
    \includegraphics[width=0.20\textwidth]{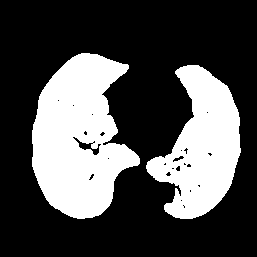} &
    \includegraphics[width=0.20\textwidth]{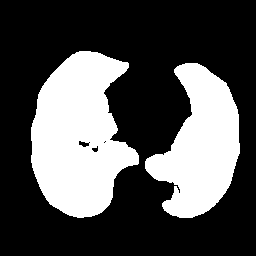} &
    \includegraphics[width=0.3cm, height=3.5cm]{Figures/dark_patch.jpg} &
    \includegraphics[width=0.20\linewidth]{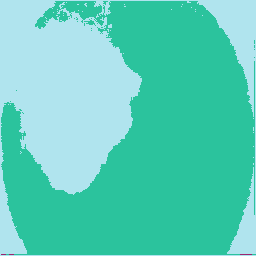} &
    \includegraphics[width=0.20\linewidth]{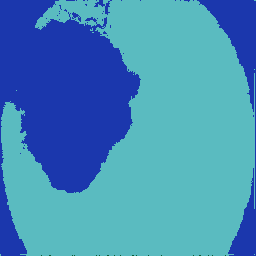} &
    \includegraphics[width=0.20\linewidth]{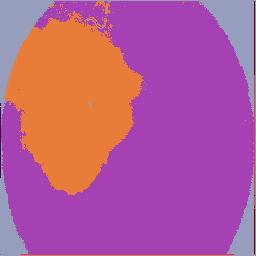} &
    \includegraphics[width=0.20\linewidth]{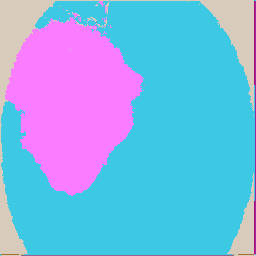} \\
    {\Large Input Image} & {\Large Ground Truth} & {\Large \textit{k}-means} & {\Large DeepCluster} & {\Large SGSCN} & {\Large MS-Former} & {\Large Our Method} & & {\Large $\mathcal{L}_{ce}$} & {\Large $\mathcal{L}_{ce} + \mathcal{L}_{B}$} & {\Large $\mathcal{L}_{ce} + \mathcal{L}_{CLIP}$} & {\Large $\mathcal{L}_{joint}$}
    \end{tabular}
    } 
    \vspace{-0.75em}
    \caption{Visual comparison of different methods on the PH$^2$ skin lesion segmentation and Lung datasets (left) and the impact of individual loss functions (right).} 
    \vspace{-1em}
    \label{fig:skin}    
\end{figure*}

\begin{figure}[t]
    \centering
    \resizebox{0.85\linewidth}{!}{
    \begin{tabular}{@{} *{4}c @{}}
    \includegraphics[width=0.35\linewidth]{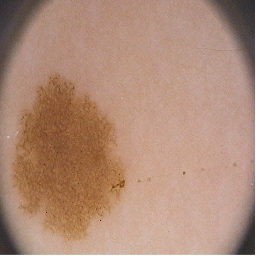} &
    \includegraphics[width=0.35\linewidth]{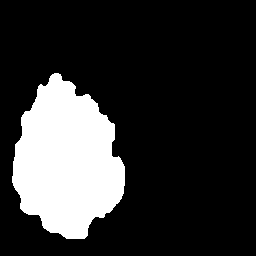} &
    \includegraphics[width=0.35\linewidth]{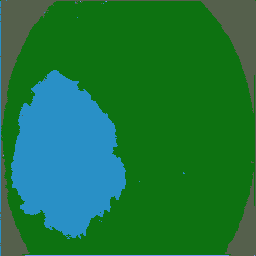} &
    \includegraphics[width=0.35\linewidth]{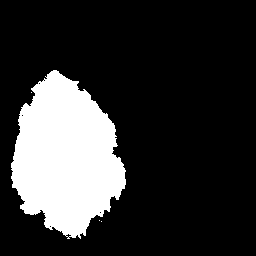}
    \\
    {Input Image} & {Ground Truth} & {Output} & {Output Mask} 
    \\
    \includegraphics[width=0.35\linewidth]{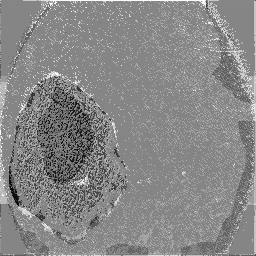} &
    \includegraphics[width=0.35\linewidth]{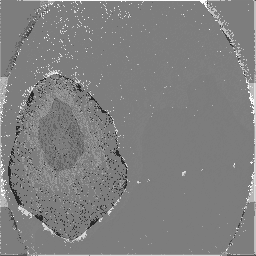} &
    \includegraphics[width=0.35\linewidth]{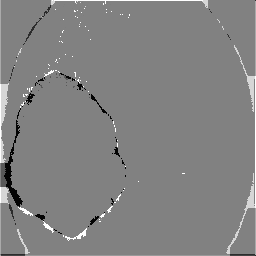} &
    \includegraphics[width=0.35\linewidth]{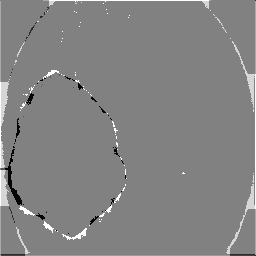} \\
     {Iter 1} & {Iter 2} & {Iter 25} & {Iter 50}
    \end{tabular}
    } 
    \vspace{-0.5em}
    \caption{Illustrating Edge Refinement Loss Impact: Visualizations show how this module enhances boundary information, aiding effective object separation during training.} 
    \vspace{-1.25em}
    \label{fig:boundary}
\end{figure}
\vspace{-1em}
\section{Experiments}
\vspace{-0.75em}
\subsection{Experimental Setup}
\vspace{-0.5em}
\noindent\textbf{Dataset}: First, we followed the same strategy outlined in~\cite{ahn2021spatial} and utilized the PH2 dataset, introduced by Mendonça et al.~\cite{mendoncca2013ph}, which comprises 200 RGB images of melanocytic lesions. This dataset encompasses a wide range of lesion types, presenting a challenging real-world problem. We used all 200 samples in our evaluation. Second, we segmented lungs within CT images using the publicly available lung analysis dataset from Kaggle, described in~\cite{azad2019bi}, which includes 2D and 3D CT images. We followed the dataset preparation and evaluation approach outlined in~\cite{karimijafarbigloo2023ms}.

\noindent\textbf{Evaluation Methodology}:
To assess our approach, we use a set of evaluation metrics, including the Dice (DSC) score, XOR metric, and Hammoud distance (HM). These metrics allow us to comprehensively compare our method to two benchmark techniques: the unsupervised \textit{k}-means clustering method and recent self-supervised strategies, specifically DeepCluster~\cite{caron2018deep}, IIC~\cite{ji2019invariant}, spatial guided self-supervised strategy (SGSCN)~\cite{ahn2021spatial}, and MS-Former~\cite{karimijafarbigloo2023ms}.
Following~\cite{ahn2021spatial,karimijafarbigloo2023ms,gharawi2023self}, we only consider the cluster with the highest overlap with the ground truth map as the target class prediction when evaluating our method. In addition, we optimize the weighting factors $\lambda_1$, $\lambda_2$, and $\lambda_3$ using~\cite{optuna_2019} and set them to 2.5, 0.5, and 0.5, respectively for both datasets. We set $\alpha$ to 3 and the downsampling factor $\beta$ to 16.
\vspace{-1em}
\subsection{Results}
\vspace{-1em}
\begin{table}[h]
    \caption{The performance of the proposed method is compared to the SOTA approaches on the PH$^2$ and Lung datasets.}\vspace{-0.5em}\label{tab:results}
    \centering
    \resizebox{0.49\textwidth}{!}{
    \arrayrulecolor{black}
    \begin{tabular}{c||ccc||ccc} 
    \toprule
    \multirow{2}{*}{\textbf{Methods}} & \multicolumn{3}{c||}{\textbf{PH$^2$}} & \multicolumn{3}{c}{\textbf{Lung Segmentation}} \\ 
    \cline{2-7}
     & \textbf{DSC~$\uparrow$} & \textbf{HM~$\downarrow$} & \textbf{XOR~$\downarrow$} & \textbf{DSC~$\uparrow$} & \textbf{HM~$\downarrow$} & \textbf{XOR~$\downarrow$} \\ 
    \hline
    \textit{k}-means & 71.3 & 130.8 & 41.3 & 92.7 & 10.6 & 12.6 \\
    DeepCluster~\cite{caron2018deep} & 79.6 & 35.8 & 31.3 & 87.5 & 16.1 & 18.8 \\
    IIC~\cite{ji2019invariant} & 81.2 & 35.3 & 29.8 & - & - & - \\
    SGSCN\cite{ahn2021spatial} & 83.4 & 32.3 & 28.2 & 89.1 & 16.1 & 34.3 \\ 
    MS-Former~\cite{karimijafarbigloo2023ms} & 86.0 & 23.1 & 25.9 & 94.6 & 8.1 & 14.8\\
    \hline
    \rowcolor[rgb]{1,0.917,0.776}
    \textbf{$\mathcal{L}_{ce}$} & 86.7 & 22.1 & 24.7 & 93.2 & 9.4 & 5.7\\
    \rowcolor[rgb]{1,0.917,0.776}
    \textbf{$\mathcal{L}_{ce} + \mathcal{L}_{B}$} &  87.8 & 20.3 & 21.5 & 93.6 & 9.0 & 5.4 \\
    \rowcolor[rgb]{1,0.917,0.776}
    \textbf{$\mathcal{L}_{ce} + \mathcal{L}_{CLIP}$} & 87.9 & 20.2 & 21.6 & 93.7 & 9.1 & 5.6 \\
    \hline
    \rowcolor[rgb]{1,0.772,0.36}
    \textbf{Our Method} & \textbf{88.7} & \textbf{19.3} & \textbf{20.1} & \textbf{95.3} & \textbf{7.2} & \textbf{4.7}\\
    \bottomrule
    \end{tabular}
    }
    \vspace{-0.5em}
\end{table}

In Table~\autoref{tab:results}, we present the segmentation results for both the PH$^2$ and Lung organ segmentation datasets. Our method, FuseNet, achieves superior performance on both skin lesion and lung segmentation tasks, as evidenced by the higher DSC scores and lower HM and XOR values. Notably, FuseNet outperforms SGSCN and MS-Former by leveraging several key components. First, we use the CLIP method to model the consistency between two views of the image, harnessing contextual information. Additionally, we introduce an edge refinement loss function that minimizes the disparity between the segmentation map and its downsampled and then upsampled counterpart. This process generates an edge map, which is crucial for separating overlapped boundaries, especially in the case of skin lesions with deformable shapes. Our dual-stream method, guided by CLIP, is adept at modeling local texture details and global context dependencies among image views. This improves clustering, as shown in~\autoref{tab:results}. Going beyond quantitative metrics, we also present qualitative results.~\autoref{fig:skin} illustrates the visual segmentation of both datasets, highlighting the effectiveness of our model in improving segmentation by increasing the number of true positives and reducing the number of false positives. 

\vspace{-1.25em}
\section{Ablation Study}
\vspace{-1em}
To assess the individual impact of the CLIP module and the spatial loss function within our architecture, we conducted a systematic experimental analysis by selectively deactivating these components (see \autoref{tab:results}). Our results show that a modest 0.9\% reduction in the DSC score was observed in the PH$^2$ dataset when the CLIP module was excluded, underscoring its significant contribution to segmentation accuracy. Similarly, removing the edge refinement loss function resulted in a 0.8\% decline in DSC performance, emphasizing its crucial role in maintaining spatial coherence. These findings are visually presented in~\autoref{fig:skin}, illustrating the consequences of excluding these modules on segmentation results. To further elucidate the influence of our edge refinement loss, we provide visualizations of edge information throughout the training process, demonstrating how this module enhances boundary information, ultimately facilitating effective object separation (see \autoref{fig:boundary}).

\vspace{-1.25em}
\section{Conclusion}
\vspace{-1em}
FuseNet excels in challenging medical image segmentation scenarios, substantially improving segmentation quality. It outperforms SOTA methods based on DSC score, HM, and XOR metrics. Visual results highlight FuseNet's ability to increase true positives and reduce false positives, advancing self-supervised medical image analysis and reducing the need for expensive  manual annotations.

\bibliographystyle{IEEEbib}
\bibliography{refs.bib}

\begin{thebibliography}{10}

\bibitem{azad2022medical}
Reza Azad, Ehsan~Khodapanah Aghdam, Amelie Rauland, Yiwei Jia, Atlas~Haddadi Avval, Afshin Bozorgpour, Sanaz Karimijafarbigloo, Joseph~Paul Cohen, Ehsan Adeli, and Dorit Merhof,
\newblock ``Medical image segmentation review: The success of u-net,''
\newblock {\em arXiv preprint arXiv:2211.14830}, 2022.

\bibitem{gharawi2023self}
Abdulrahman Gharawi, Mohammad~D Alahmadi, and Lakshmish Ramaswamy,
\newblock ``Self-supervised skin lesion segmentation: An annotation-free approach,''
\newblock {\em Mathematics}, vol. 11, no. 18, pp. 3805, 2023.

\bibitem{karimijafarbigloo2023ms}
Sanaz Karimijafarbigloo, Reza Azad, Amirhossein Kazerouni, and Dorit Merhof,
\newblock ``Ms-former: Multi-scale self-guided transformer for medical image segmentation,''
\newblock in {\em Medical Imaging with Deep Learning}, 2023.

\bibitem{karimijafarbigloo2023self}
Sanaz Karimijafarbigloo, Reza Azad, Amirhossein Kazerouni, Yury Velichko, Ulas Bagci, and Dorit Merhof,
\newblock ``Self-supervised semantic segmentation: Consistency over transformation,''
\newblock in {\em Proceedings of the IEEE/CVF International Conference on Computer Vision}, 2023, pp. 2654--2663.

\bibitem{ahn2021spatial}
Euijoon Ahn, Dagan Feng, and Jinman Kim,
\newblock ``A spatial guided self-supervised clustering network for medical image segmentation,''
\newblock in {\em Medical Image Computing and Computer Assisted Intervention--MICCAI 2021: 24th International Conference, Strasbourg, France, September 27--October 1, 2021, Proceedings, Part I 24}. Springer, 2021, pp. 379--388.

\bibitem{taher2022caid}
Mohammad Reza~Hosseinzadeh Taher, Fatemeh Haghighi, Michael~B Gotway, and Jianming Liang,
\newblock ``Caid: Context-aware instance discrimination for self-supervised learning in medical imaging,''
\newblock in {\em International Conference on Medical Imaging with Deep Learning}. PMLR, 2022, pp. 535--551.

\bibitem{he2023geometric}
Yuting He, Guanyu Yang, Rongjun Ge, Yang Chen, Jean-Louis Coatrieux, Boyu Wang, and Shuo Li,
\newblock ``Geometric visual similarity learning in 3d medical image self-supervised pre-training,''
\newblock in {\em Proceedings of the IEEE/CVF Conference on Computer Vision and Pattern Recognition}, 2023, pp. 9538--9547.

\bibitem{huang2021missformer}
Xiaohong Huang, Zhifang Deng, Dandan Li, and Xueguang Yuan,
\newblock ``Missformer: An effective medical image segmentation transformer,''
\newblock {\em arXiv preprint arXiv:2109.07162}, 2021.

\bibitem{radford2021learning}
Alec Radford, Jong~Wook Kim, Chris Hallacy, Aditya Ramesh, Gabriel Goh, Sandhini Agarwal, Girish Sastry, Amanda Askell, Pamela Mishkin, Jack Clark, et~al.,
\newblock ``Learning transferable visual models from natural language supervision,''
\newblock in {\em International conference on machine learning}. PMLR, 2021, pp. 8748--8763.

\bibitem{mendoncca2013ph}
Teresa Mendon{\c{c}}a, Pedro~M Ferreira, Jorge~S Marques, Andr{\'e}~RS Marcal, and Jorge Rozeira,
\newblock ``Ph 2-a dermoscopic image database for research and benchmarking,''
\newblock in {\em 2013 35th annual international conference of the IEEE engineering in medicine and biology society (EMBC)}. IEEE, 2013, pp. 5437--5440.

\bibitem{azad2019bi}
Reza Azad, Maryam Asadi-Aghbolaghi, Mahmood Fathy, and Sergio Escalera,
\newblock ``Bi-directional convlstm u-net with densley connected convolutions,''
\newblock in {\em ICCV 2019, IEEE International Conference on Computer Vision 2019}, 2019.

\bibitem{caron2018deep}
Mathilde Caron, Piotr Bojanowski, Armand Joulin, and Matthijs Douze,
\newblock ``Deep clustering for unsupervised learning of visual features,''
\newblock in {\em Proceedings of the European conference on computer vision (ECCV)}, 2018, pp. 132--149.

\bibitem{ji2019invariant}
Xu~Ji, Joao~F Henriques, and Andrea Vedaldi,
\newblock ``Invariant information clustering for unsupervised image classification and segmentation,''
\newblock in {\em Proceedings of the IEEE/CVF International Conference on Computer Vision}, 2019, pp. 9865--9874.

\bibitem{optuna_2019}
Takuya Akiba, Shotaro Sano, Toshihiko Yanase, Takeru Ohta, and Masanori Koyama,
\newblock ``Optuna: A next-generation hyperparameter optimization framework,''
\newblock in {\em Proceedings of the 25th {ACM} {SIGKDD} International Conference on Knowledge Discovery and Data Mining}, 2019.

\end{thebibliography}

\end{document}